\documentclass{article}

\usepackage{arxiv}
\usepackage{hyperref}       
\usepackage[utf8]{inputenc} 
\usepackage[T1]{fontenc}    
\usepackage{url}            
\usepackage{booktabs}       
\usepackage{amsfonts}       
\usepackage{nicefrac}       
\usepackage{microtype}      
\usepackage{lipsum}		
\usepackage{graphicx}
\usepackage[numbers]{natbib}

\usepackage{doi}

\usepackage{amsfonts}       
\usepackage{amsmath, amssymb, bm}

\usepackage{comment}
\graphicspath{ {./figures/} }

\usepackage{placeins}

\usepackage{color}
\newcommand\changes[1]{\textcolor{black}{#1}}

\title{Enhancing Mechanical Metamodels with a Generative Model-Based Augmented Training Dataset}

\author{ \href{https://orcid.org/
0000-0002-8404-1429}{\includegraphics[scale=0.06]{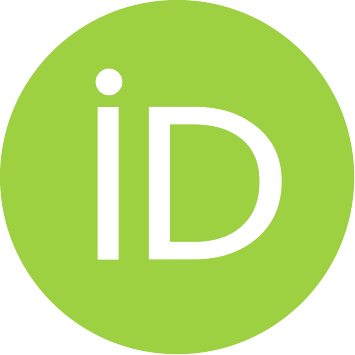}\hspace{1mm}Hiba Kobeissi}\thanks{co-first authors} \\
	Department of Mechanical Engineering\\
	Boston University\\
    Boston, MA 02215 \\
	\texttt{hibakob@bu.edu} \\
	\And
	\href{https://orcid.org/0000-0001-9879-044X}{\includegraphics[scale=0.06]{orcid.pdf}\hspace{1mm}Saeed Mohammadzadeh$^*$} \\
	Department of Systems Engineering\\
	Boston University\\
	Boston, MA 0215 \\
	\texttt{saeedmhz@bu.edu} \\
	\And
	\href{https://orcid.org/0000-0001-8099-3468}{\includegraphics[scale=0.06]{orcid.pdf}\hspace{1mm}Emma Lejeune}\thanks{corresponding author} \\
	Department of Mechanical Engineering\\
	Boston University\\
    Boston, MA 02215 \\
	\texttt{elejeune@bu.edu} \\
}

\renewcommand{\shorttitle}{Enhancing Mechanical Metamodels with Generative Model-Based Augmentation}

\hypersetup{
pdftitle={Enhancing Mechanical Metamodels with a Generative Approach},
pdfsubject={q-bio.NC, q-bio.QM},
pdfauthor={Hiba Kobeissi, Saeed Mohammadzadeh, Emma Lejeune},
pdfkeywords={machine learning, mechanics, surrogate modeling},
}

\begin{document}
\maketitle

\begin{abstract}

\changes{Modeling biological soft tissue is complex in part due to material heterogeneity. Microstructural patterns, which play a major role in defining the mechanical behavior of these tissues, are both challenging to characterize, and difficult to simulate. Recently, machine learning (ML)-based methods to predict the mechanical behavior of heterogeneous materials have made it possible to more thoroughly explore the massive input parameter space associated with heterogeneous blocks of material. Specifically, we can train ML models to closely approximate computationally expensive heterogeneous material simulations where the ML model is trained on datasets of simulations with relevant spatial heterogeneity. However, when it comes to applying these techniques to tissue, there is a major limitation: the number of useful examples available to characterize the input domain under study is often limited. In this work, we investigate the efficacy of both ML-based generative models and procedural methods as tools for augmenting limited input pattern datasets. We find that a Style-based Generative Adversarial Network with an adaptive discriminator augmentation mechanism is able to successfully leverage just 1,000 example patterns to create authentic generated patterns. And, we find that diverse generated patterns with adequate resemblance to real patterns can be used as inputs to finite element simulations to meaningfully augment the training dataset. To enable this methodological contribution, we have created an open access Finite Element Analysis simulation dataset based on Cahn-Hilliard patterns. We anticipate that future researchers will be able to leverage this dataset and build on the work presented here.}

\end{abstract}

\keywords{machine learning \and mechanics \and surrogate modeling \and heterogeneous materials}

\section{Introduction}
\label{sec:intro}

Establishing models that realistically capture the biomechanical behavior of soft tissue is a challenging yet crucial endeavor \citep{Barocas2021hybrid, Fan2014Simulation}. High fidelity mechanical models are needed for tasks such as surgical simulation \citep{Berkley2004Real, joldes2009suite, Zhang2018Deformable}, patient-specific procedure planning \citep{Joldes2019Suite, SahliCostabal2020Classifying}, modeling of in-vivo biological mechanisms \citep{Sree2019linking, kong2018finite}, and inverse material characterization \citep{Kakaletsis2021Right, avazmohammadi2018integrated}. Capturing the mechanical behavior of soft tissue is challenging because soft tissues are often highly nonlinear and anisotropic, they can exhibit a nonlinear stiffening response, they often undergo large deformations, and they have a complex hierarchical structure \citep{Barocas2021hybrid, Ogden2017, Ateshian2017, Holzapfel2001Biomechanics}. For example, at the microstructural level, soft tissue may contain components such as fibers with a preferred direction which give rise to highly anisotropic material behavior on the macroscale \citep{Holzapfel2001Biomechanics}. In addition to complex constitutive behavior, biological materials are also challenging to model because they tend to be highly heterogeneous \citep{Ateshian2017, Humphrey2013Multiscale}. As such, developing faithful mechanical models of soft tissues and numerically implementing them (e.g., in the Finite Element setting \citep{Hughes2012Finite}) is both challenging and typically quite computationally expensive \citep{Fan2014Simulation, Kakaletsis2021Right, Holzapfel2001Biomechanics,Barocas2019Comp,Gomez2021Poroelasticity,Thao2021general}. Notably, both the exact values of the mechanical properties of biological tissue and their heterogeneous distribution in space are often uncertain \citep{Madireddy2015Bayesian, Jadidi2021Constitutive}. Therefore, in order to get a true picture of tissue behavior, it is necessary to run multiple simulations that capture the range of relevant input parameters \citep{Kakaletsis2021Right}. 
In this context, there has been substantial recent interest in reducing the computational cost of these numerical simulations at the cost of marginal decrease in the simulation accuracy \citep{tonutti2017machine}. 

\changes{Markedly}, there has been recent interest in using machine learning tools to create computationally inexpensive data-driven models of soft biological tissue in particular \citep{tac2021datadriven}, and for various biomedical applications in general \citep{Yongjie2021imagebased,Yongjie2021deep,peng2021multiscale,peirlinck2019using}.
In previous work by our group and others \citep{lejeune2021exploring, lejeune2020mechanical, mohammadzadeh2022predicting, Yang2021Deep, mianroodi2021teaching, Stowers2021Improving, yang2019derivation, Liu2019Deep}, metamodels, or surrogate models \citep{forrester2009recent}, developed with supervised machine learning algorithms and multi-fidelity mechanical datasets have been used successfully to predict the mechanical behavior of heterogeneous materials via single and full-field Quantities of Interest (QoIs) (e.g. strain energy, displacement/strain fields, damage fields). 
For example, Tonutti et al. \citep{tonutti2017machine} used the results of Finite Element Analysis (FEA) simulations in conjunction with artificial neural networks and support vector regression to develop computationally inexpensive patient-specific deformation models for brain pathologies. In addition, Salehi et al. \citep{salehi2021physgnn} trained graph neural networks with FEA simulation results to speed-up the approximation of soft tissue deformation with acceptable loss of accuracy for neurosurgical applications. And, in Tac et al. \citep{tac2021datadriven}, fully connected neural networks were trained with high-fidelity biaxial test data and low-fidelity analytical approximations to derive a data-driven anisotropic constitutive model of porcine and murine skin. Notably, due to the limited availability of both experimental data and high fidelity simulation data, methods that rely on multiple data fidelities (i.e., multi-fidelity models) have been shown to be more effective than single fidelity schemes given a \changes{small number} of high fidelity data \citep{tac2021datadriven,lejeune2021exploring, lu2020extraction, teichert2019machine}. 
This is particularly true for methods that rely on deep learning where training datasets must be large for successful model implementation \citep{Gomez2021PhysicsInformed,brock2019large,karras2020training}.
Though multi-fidelity methods can address the scenario where there are limited high-fidelity simulations results, they are not necessarily equipped to address the scenario where there is limited information about what the training dataset should contain. 
For example, it is unlikely that researchers will have tens of thousands of accurate examples of the heterogeneous material property distribution of a given soft tissue of interest. 
In this work, our goal is to systematically answer the question: is it possible to create a meaningful training dataset \changes{that ultimately improves the performance of} a deep learning-based metamodel of heterogeneous material given only a small number of representative examples of the relevant material property distribution input pattern? 

To address this question, we first define a benchmark problem to evaluate our proposed machine learning approach. This is important because, at present, there are only a small number of existing open access benchmark datasets related to problems in solid mechanics \citep{lejeune2020mechanicalF,lejeune2019mechanicalUE,mohammadzadeh2021mechanical,pprachas2022Asymmetric}. Furthermore, of the available datasets, few contain a good representation of the heterogeneous material properties most relevant to soft tissue modeling. 
Our benchmark dataset, the ``Mechanical MNIST Cahn-Hilliard'' dataset, is a contribution to our previously initiated ``Mechanical MNIST'' project where we provide simulation results for heterogeneous materials undergoing large deformation. 
The full dataset contains $104,813$ Cahn-Hilliard patterns and associated equibiaxial extension simulations, and it is straightforward to train a deep learning-based metamodel to predict QoI from these simulations (e.g., change in strain energy $\Delta \Psi$). 
However, if we constrain ourselves to only a small subset of these example input patterns, for example, if we limit our knowledge to just $1\text{,}000$ example patterns, it becomes much more challenging to effectively train a deep learning-based metamodel. 
With this benchmark dataset and imposed limitation, we are able to test \changes{both} the efficacy of \changes{ML-based} generative models, models that learn the data distribution and generate plausible examples from the distribution \citep{Gansgoodfellow}, \changes{and procedural methods} at augmenting a constrained version of the available training dataset. By comparing the results of metamodels that rely on generated patterns to metamodels that are trained on true input patterns, we are able to systematically evaluate the efficacy of our proposed \changes{size-limited data augmentation approaches.} 
We note that this premise follows from recent work in the literature where generative models have been used to augment small materials characterization datasets  \citep{Jangid2022Grain,ma2020image}. Ultimately, we are able to clearly demonstrate that leveraging the capabilities of our selected \changes{data generation models} is an effective tool for augmenting small datasets of material property distributions in biological tissue for the purpose of creating training datasets for \changes{ML}-based metamodels.  

The remainder of the paper is organized as follows. In Section \ref{sec:meth}, we begin by introducing our ``Mechanical MNIST Cahn-Hilliard'' dataset. Then, we describe our approach to training a metamodel to approximate the mechanical behavior of the simulations, and our approach to generating synthetic input patterns to augment the training dataset. In Section \ref{sec:res_disc}, we show the performance of our generative model\changes{s}, and the performance of our metamodel with \changes{ML-based and procedural} augmented training dataset. We conclude in Section \ref{sec:conc}. Finally, we note briefly that links to the code and dataset required to reproduce our work are given in Section \ref{sec:additional_info}.

\section{Methods}
\label{sec:meth}

Here, we begin in Section \ref{sec:meth_dataset} with an introduction to our ``Mechanical MNIST Cahn-Hilliard'' dataset. Then, in Section \ref{sec:meth_simple_metamodel}, we \changes{describe} our metamodeling approach where a \changes{ML}-based metamodel is used to predict a single quantity of interest (in this case change in strain energy $\Delta \Psi$) from an array-based representation of the input pattern. \changes{Then,} in Section \ref{sec:meth_gen_model}, we detail our \changes{three different approaches} to ML-based generative modeling of the input pattern distribution. \changes{In Section \ref{sec:rbPatterns}, we introduce two additional procedural methods for generating synthetic input patterns. And in Section \ref{sec:evalM}, we present the evaluation metrics that we considered to compare the performance of the different methods that we have implemented to generate synthetic patterns. Finally, in \ref{sec:rotation} we define our procedure for standard rotation-based augmentation. We briefly note that in order to stay consistent with the literature, the Greek letters $\lambda$ and $\mu$ refer to different constants in Sections \ref{sec:meth_dataset} and \ref{sec:evalM}. In both cases, we provide a brief definition of each term when it is introduced.}

\subsection{The Mechanical MNIST Cahn-Hilliard Dataset}
\label{sec:meth_dataset}

In conjunction with our previous publications \citep{lejeune2021exploring, lejeune2020mechanical,mohammadzadeh2022predicting}, we introduced the ``Mechanical MNIST'' dataset of heterogeneous materials undergoing large deformation. In previous iterations of the dataset, heterogeneous input domain patterns were defined by the MNIST \citep{mnist} and Fashion MNIST \citep{fashion-mnist} bitmap patterns. 
For this manuscript, we extend our ``Mechanical MNIST'' dataset collection to include additional patterns from a different input domain distribution that is more relevant to heterogeneous biological materials. 
\changes{The input patterns for the ``Mechanical MNIST Cahn-Hilliard'' dataset are generated based on} Alan Turing's model of morphogenesis \citep{turing1990chemical} -- a common motif during biological development manifested in many different animal and plant patterns such as the pigmentation of animal skins, the branching of trees and other skeletal structures, and the distinct patterns on leaves and petals \citep{maini2012turing,garikipati2017perspectives}. We obtain these patterns by solving a nonlinear spatio-temporal fourth-order partial differential equation (PDE) referred to as the Cahn-Hilliard equation, that was originally proposed to describe the process of phase separation in isotropic binary alloys \citep{grant1993spinodal,wang2019,wang2021}.

\begin{figure}[h]
\begin{center}
\includegraphics[width=.8\textwidth]{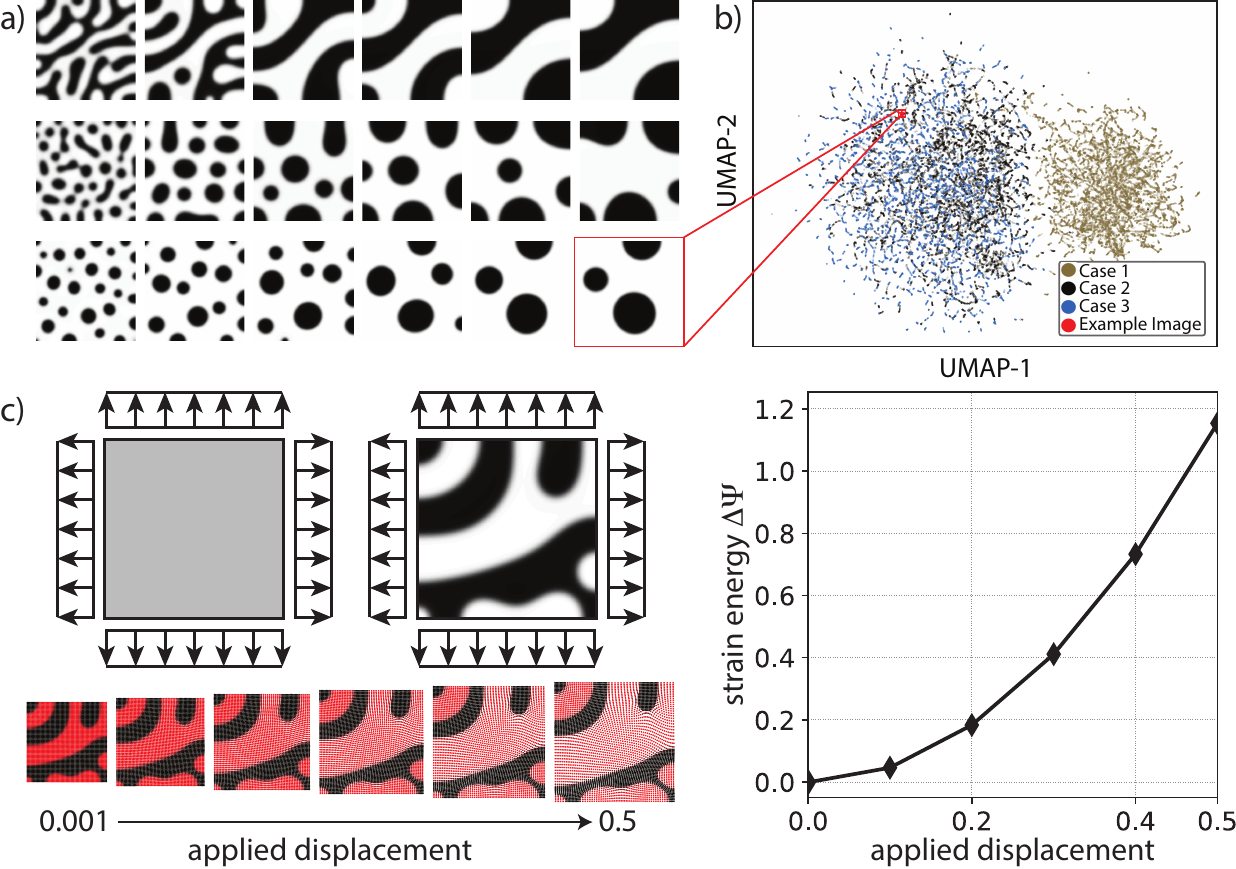}
\caption{\label{fig:dataset}a) Illustration of the spatial patterns obtained from our Cahn-Hilliard simulations where each row corresponds to the time evolution in a single simulation for $c_0 = 0.5$ (Case $1$), $c_0 = 0.63$ (Case $2$) and $c_0 = 0.75$ (Case $3$) shown in the first, second and the third rows respectively. b) A UMAP visualization \citep{sainburg2021parametric} of a representative proportion of our Cahn-Hilliard patterns  using {\fontfamily{pcr}\selectfont random\_state=42, n\_neighbors=30, min\_dist=0.1} as training parameters. c) A schematic illustration of displacement driven equibiaxial extension applied to a heterogeneous domain dictated by the Cahn-Hilliard patterns. Here we show an example from Case $1$: $c_0=0.5$ and plot the deformed state at the six magnitudes of applied displacement. From these Finite Element simulations, we obtain multiple outputs including the total change in strain energy $\Delta \Psi$ at each load step.}
\end{center}
\end{figure}

Our new dataset, ``Mechanical MNIST Cahn-Hilliard'', contains not only Cahn-Hilliard based two-dimensional heterogeneous input patterns, but also the results of Finite Element simulations of these material domains subjected to equibiaxial extension. Here we will summarize the process of creating this dataset. 
Briefly, the Cahn-Hilliard equation, which is a fourth-order partial differential equation that governs the evolution of a binary mixture, can first be reduced to a pair of second-order equations \citep{wells2006discontinuous,FEniCSdemo}. This mixed formulation can be expressed in the weak form for the two unknown fields, $c$, the concentration of one of the components of the binary mixture, and $\mu$, the chemical potential of a uniform solution:

\begin{align}
\int_\Omega \frac{c_{n+1}-c_n}{t_{n+1}-t_n}q \, dx+\int_\Omega M\nabla \mu_{n+\theta} \cdot \nabla q \, dx=0\qquad &\forall \, q\in V \label{weak1}\\
\int_\Omega \mu_{n+1} v \, dx-\int_\Omega \frac{df_{n+1}}{dc} v \, dx-\int_\Omega \lambda \nabla c_{n+1} \cdot \nabla v \, dx =0\qquad &\forall \, v\in V \label{weak2}
\end{align}

where $M$ is the mobility parameter, $\lambda$ is a positive scalar that describes the thickness of the interfaces between the phases of the mixture, $f$ is the chemical free-energy function, and $q$ and $v$ are test functions \citep{wells2006discontinuous, FEniCSdemo}. 

We solve the Cahn-Hilliard equations using the open source Finite Element software FEniCS \citep{alnaes2015fenics,logg2012automated} and run $2\text{,}072$ phase separation simulations on a unit square domain $\Omega = [0,1]$ where each simulation differs in the following: (1) the initial concentration $c_0$ with uniform random fluctuations of zero mean and range between $-0.05$ and $0.05$, (2) the grid size on which the initialized concentration is allowed to spatially vary, (3) the interface thickness $\lambda$, and (4) the peak-to-valley value of the free-energy function $f$, a symmetric double-well function. We record the concentration parameter at multiple time steps in each simulation to obtain $105\text{,}427$ spatial distribution patterns which broadly fall under two qualitative types: spotted (for $c_0 = 0.63$ and $c_0 = 0.75$), and striped (for $c_0 = 0.5$)\changes{, as is expected for these types of simulations \citep{wells2006discontinuous,wodo2011computationally,gomez2008isogeometric}, and store the obtained images as $400\times400$ binary bitmaps.} Example patterns are illustrated in Fig. \ref{fig:dataset}a. For further details on the underlying theory of the Cahn-Hilliard equation and our Finite Element implementation, we refer the reader to the supplementary document provided with the dataset (see Section \ref{sec:additional_info}). As an additional step, we visualize downsampled $64 \times 64$ vectors describing each Cahn-Hilliard pattern array in a two-dimensional space using the dimension reduction technique\changes{,} Uniform Manifold Approximation and Projection (UMAP) \citep{sainburg2021parametric}\changes{, which provides us with a qualitative tool to visualize our high-dimensional dataset input parameter space. Notably,} the plot in Fig.\ref{fig:dataset}b clearly reveals the two distinct clusters of patterns \changes{which is consistent with our observation that the dataset is split between the striped and spotted motifs.}

From this collection of $105\text{,}427$ heterogeneous input patterns, we perform a second set of Finite Element simulations where we use the input patterns to inform the heterogeneous material property distribution of the domain and subject it to equibiaxial extension. To accomplish this, we first convert the binary bitmap patterns into meshed domains of two different materials. \changes{Briefly, we detect the contours of the image features and extract their coordinates using the OpenCV library \citep{bradski2008learning}. We then translate these coordinates into a mesh with two different subdomains, background and pattern, using pygmsh $6.1.1$ \citep{nico_schlomer_2020_3764683}, a Python implementation of Gmsh $4.6.0$ \citep{geuzaine2009gmsh}.} We note briefly that from our initial collection of $105\text{,}427$ images, $614$ images could not be processed because they exhibited either pattern features that were too small to be detected as area domains, features that were in very close proximity to each other, or complex hierarchical contours that our pipeline was not able to detect and process. Thus, our final dataset contains $104\text{,}813$ simulation results. 
Based on a mesh refinement study, we chose quadratic triangular elements with a characteristic length of $0.01$. This led to approximately $41\text{,}000$ elements in a typical domain.

Once the material domain was meshed, we performed equibiaxial extension simulations in FEniCs \citep{alnaes2015fenics,logg2012automated}. Here we chose a compressible Neo-Hookean material model defined by strain energy $\Psi$ as:
\begin{equation}
\Psi=\frac{1}{2} \, \mu \,  \big[\textbf{F}:\textbf{F}-3-2\ln(\det\textbf{F})\big]+\frac{1}{2} \, \lambda \, \big[\frac{1}{2}[(\det\textbf{F})^{2}-1]-\ln(\det\textbf{F})\big]
\end{equation}
where $\textbf{F}$ is the deformation gradient, and $\mu$ and $\lambda$ are the Lam{\'e} parameters equivalent to Young's modulus $E$ and Poisson's ratio $\nu$ as $E = \mu \,  (3\lambda+2\mu)/(\lambda+\mu)$ and $\nu=\lambda/(2(\lambda+\mu))$. 
We define the Poisson's ratio as a constant $(\nu = 0.3)$, and we specify a Young's modulus $E$ for the background domain that is 10 times lower than the Young's modulus for the ``stiffer'' spotted and striped patterns $(E = [1, 10])$. We set up each Finite Element simulation for equibiaxial deformation so that every external edge of the domain is extended by half of the value of given applied displacement in the direction of the outward normal to the surface (Fig.\ref{fig:dataset}c). The set of fixed displacements \textbf{d} go up to $50\%$ of the initial domain size as: 
\begin{equation}
\textbf{d}=[0.0,0.001,0.1,0.2,0.3,0.4,0.5] \, .
\end{equation}
The output of each of the $104\text{,}813$ large deformation simulations consisted of data on the total change in strain energy $\Delta \Psi$, total reaction force in the $x$ and $y$ directions, and full field domain displacement collected on a downsampled $64\times64$ grid (Fig.\ref{fig:dataset}\changes{c}). We chose the size of the grid to be the smallest possible size that could be reached without the loss of important image features. \changes{In this context, we consider the borders of the white/dark patterns to be important features that should not be distorted much by any operation to avoid misclassifiying the cells along the edges into the wrong subdomain.}
We note that all code to implement these simulations is shared on GitHub with access details given in Section \ref{sec:additional_info}.

\subsection{Metamodel Design and Implementation}
\label{sec:meth_simple_metamodel}

In this Section, we summarize our approach to creating metamodels for predicting the change in strain energy $\Delta \Psi$ from the input Cahn-Hilliard patterns. \changes{In Sections \ref{sec:meth_gen_model}, \ref{sec:rbPatterns}, and \ref{sec:rotation}, we describe the details of our generative model-based, procedural-based and standard rotation-based approaches that we implement to augment the training dataset.} 

\subsubsection{Feedforward Convolutional Neural Network}
\label{sec:basic_CNN}

In this paper, we are focused on using machine learning techniques for predicting single quantities of interest ($\Delta \Psi$) from input arrays (Cahn-Hilliard patterns). 
This goal is illustrated schematically in Fig.\ref{fig:pipeline}a.
To accomplish this, we implemented a basic feedforward convolutional neural network (CNN) consisting of a total of $9$ convolutional layers each followed by batch normalization and rectified linear unit (ReLu) activation except for the last $(9^{th})$ layer. For downsampling input images, we used max pooling after the first three convolutional layers with \textit{same} padding while \textit{valid} padding is used for the rest of the convolutional layers. Our network has a total of $3\text{,}734\text{,}625$ trainable parameters. We trained the network using the PyTorch library \citep{NEURIPS2019_9015} with a batch size of $64$ for \changes{$100$} epochs. We employ an Adam optimizer \citep{kingma2017adam} with learning rate \changes{$\alpha = 0.01$ reduced to $0.001$ after $50$ epochs} and exponential decay rates $\beta_1 = 0.9$ and $\beta_2 = 0.999$. 
The output of the CNN is a single quantity of interest (QoI) for a $64\times64$ array input describing the simulation input pattern.
We validated our model performance through a 5-fold cross-validation approach based on Mean Squared Error (MSE). In Section \ref{sec:res_gen_perform}, we report the performance of our model on test data. 
 
\subsubsection{Transfer Learning}
\label{sec:tl}

Our original ``Mechanical MNIST Cahn-Hilliard'' dataset took approximately $5\text{,}240$ CPU hours to generate. 
Rather than expending a similar level of resources to run simulations based on generated input patterns, we decided to employ a transfer learning approach where we leverage low fidelity simulation data \citep{yosinski2014transferable}. 
Specifically, we followed the approach outlined in our recent publication \citep{lejeune2021exploring} to create low fidelity simulation versions of our dataset that are run on a coarse mesh ($64\times64$ grid, $8\text{,}192$ elements) with linear elements and only subject to a perturbation displacement ($0.001$) rather than the full $50\%$ extension. With these parameters, it took approximately \changes{$4.2$ CPU hours} to generate a low fidelity dataset of \changes{$72\text{,}000$} patterns and the corresponding strain energy values only for a perturbation displacement. 
Notably, this is \changes{$0.08\%$} of the time it would take to generate the equivalent number of high fidelity simulations \changes{described in Section \ref{sec:meth_dataset}}. Of course, this speed up comes at the price of introducing numerical error that must be subsequently dealt with through transfer learning. 

\begin{figure}[h]
\begin{center}
\includegraphics[width=.95\textwidth]{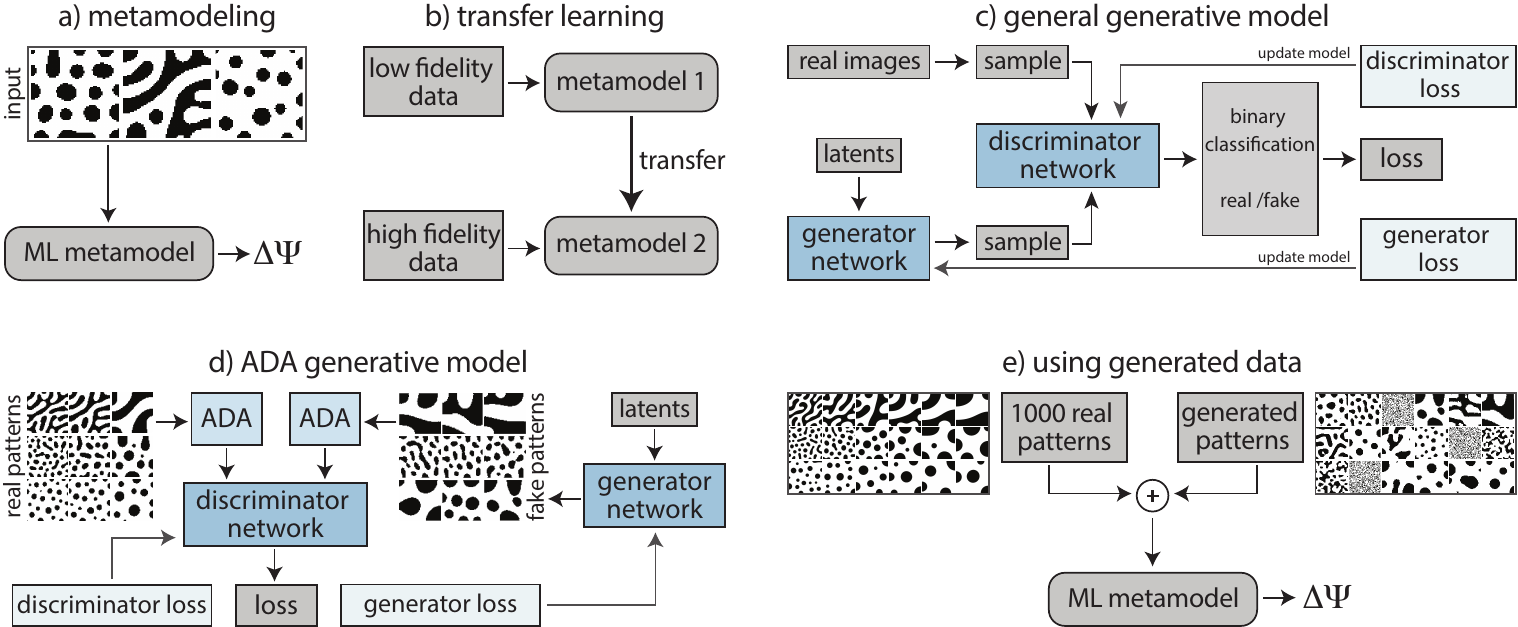}
\caption{\label{fig:pipeline}a) A schematic of our ML metamodels that are used to predict change in strain energy $\Delta \Psi$ at a fixed level of applied displacement from each material property distribution. b) A schematic of transfer learning whereby a model trained on one dataset (in this case a low fidelity dataset) is used to make predictions on another dataset (in this case a high fidelity dataset). \changes{c) Architecture of a generative adversarial network including the WGAN models trained in this paper.} d) An illustration of the StyleGAN2 with an adaptive discriminator augmentation (ADA) mechanism implemented in this work as adapted from \citep{karras2020training}. e) A schematic of combining simulations based on both generated and real patterns to create a larger training dataset.}

\end{center}
\end{figure}

Our implementation of transfer learning is a straightforward model pre-training approach illustrated schematically in Fig.\ref{fig:pipeline}b and described in detail in our previous publication \citep{lejeune2021exploring}. Part of the appeal of this approach is that it is quite straightforward to implement. First, we train the metamodel (in our case the CNN defined in Section \ref{sec:basic_CNN}) on the low fidelity dataset. Then, we use this pre-trained metamodel as the weight initialization for additional training with the high fidelity dataset. In our case, the low fidelity dataset will contain data from up to $16\text{,}000$ simulations while the high fidelity dataset will contain data from only $1\text{,}000$ simulations. The ideal outcome from this approach is to end up with a metamodel that is trained on predominantly low-fidelity data yet performs comparably to a metamodel trained on the target high fidelity dataset. In Section \ref{sec:res_gen_perform}, we first report the metamodel performance on the low fidelity dataset (Fig. \ref{fig:res_2}) and then in Section \ref{sec:res_transfer} we report the performance of \changes{the low fidelity models transferred to the high fidelity dataset via additional training with $1\text{,}000$ high fidelity real samples (Table \ref{table:res_tf})}. 

\subsection{Augmenting the Training Dataset with a \changes{ML-Based} Generative Model}
\label{sec:meth_gen_model}

The main focus of this paper is on developing techniques to effectively train the metamodels described in Section \ref{sec:meth_simple_metamodel} even when we have limited examples of the relevant input patterns needed for creating our training dataset. Here we will explore methods for leveraging limited examples of input patterns by creating synthetic input patterns from a generative model. Briefly, we implement a Style-based Generative Adversarial Network using adaptive discriminator augmentation (StyleGAN2-ADA) \citep{karras2020training}, \changes{a Wasserstein Generative Adversarial Network with weight clipping (WGAN-CP)\citep{wgan2017arjovsky}, and a \changes{WGAN} with gradient penalty (WGAN-GP) \cite{gulrajani2017WGANGP}} to generate patterns that resemble the real striped and spotted Cahn-Hilliard patterns detailed in Section \ref{sec:meth_dataset}. \changes{The architectures of the generative models explored in this work are schematically shown in Figs. \ref{fig:pipeline}c and \ref{fig:pipeline}d. We train all three GAN models} with a limited set of $1\text{,}000$ real Cahn-Hilliard patterns (the same set of real images used for training the metamodels). We then combine equibiaxial extension simulation results of both generated and real patterns to create a larger training dataset for our metamodel as shown schematically in \changes{Fig. \ref{fig:pipeline}e}. In the remainder of this Section, we provide an overview of \changes{GANs}, \changes{describe the specific GANs implemented in this work, and briefly present our alternative approaches for augmenting the metamodel training dataset via procedural pattern generation and standard rotation-based augmentation.} 

\subsubsection{Generative Adversarial Networks}
\label{sec:gans}

In the context of machine learning, generative models are models that learn data distributions such that they can then be used to output (i.e., ``generate'') plausible new examples \citep{bishop2006pattern}.  
Building upon earlier deep generative models, generative stochastic networks \citep{Bengio13deepgenerative} in particular, and inspired by the work in \citep{Gutmann10NC, Schmidhuber92predMin, Tu2007LearningGM}, Goodfellow et al. \citep{Gansgoodfellow} developed a novel framework for generative models where the generative network is put in competition with a discriminative network that learns to distinguish between a sample obtained from the real data distribution and one that is generated from the model distribution. Known as Generative Adversarial Networks (GANs), these methods consist of training two models, a generative model $G$ and a discriminative model $D$ simultaneously competing in a minimax two-player game fashion \citep{Gansgoodfellow}. In this framework, $G$ is trained to capture the input data distribution by fooling the discriminative model $D$ and maximizing the probability of the latter mistakenly labelling a sample synthesized by $G$ as one from the training data. 

In their original form, GANs have been applied to many domains including the MNIST dataset of handwritten digits \citep{MNIST-Deng, MNIST-Lecun}, the Toronto Face Database (TFD) of human faces with expressions \citep{susskind2010toronto}, and the miscellaneous CIFAR-10 dataset \citep{krizhevsky2014cifar} with promising results \citep{Gansgoodfellow}. However, major drawbacks of the method include low resolution of the generated images, relatively low variation in the output distribution, and unstable training \citep{karras2018progressive}. Furthermore, training GANs to synthesize high-quality, high-resolution output distributions typically requires at least $10^{5} \sim 10^{6}$ input images. Without a dataset of this size, the training tends to diverge as the discriminator network overfits to the small number of training data examples and can no longer provide meaningful feedback to the generator network \citep{karras2020training}. \changes{There have been many approaches to modifying the original architecture and training formulation of GANs \citep{Gansgoodfellow} to improve their performance. Alterations to the network structure such as the implementation of Deep Convolutional GANs (DCGANs) \citep{radford2016unsupervised}, where the GAN model is scaled using CNN architectures, result in more stable behavior. Other enhanced methods include Wasserstein Generative Adversarial Networks (WGANs) and Style-based Generative Adversarial Networks (StyleGANs) which are briefly described in Sections \ref{sec:wassgan} and \ref{sec:stylgan} respectively.}

\subsubsection{\changes{Wasserstein Generative Adversarial Networks}
\label{sec:wassgan}
}
\changes{In contrast to modifying the GAN network structure as in DCGANs, Wasserstein Generative Adversarial Networks (WGANs) improve the stability of GANs by replacing the bin--to--bin distance function (i.e., the \textit{Jensen--Shannon} divergence) of the original architecture with a continuous loss function, the Earth Mover (EM) or the Wasserstein-1 ($W$) distance \citep{wgan2017arjovsky}. The shortcomings of the bin--to--bin distance functions, which generally assume an alignment between the domains of the histograms being compared, are addressed by the more robust cross--bin EM distance function defined as the minimal cost of a ``transport plan'' to transform one distribution into the other \citep{rubner2000earth, rubner2001perceptual, ling2007efficient}.}

\changes{
As proposed, the original WGAN model \citep{wgan2017arjovsky} requires that the discriminator lie within a 1-Lipschitz space so that $W$ is continuous everywhere and differentiable almost everywhere. This Lipschitz constraint is enforced via weight clipping (WGAN-CP) whereby the weights of the discriminator are restricted to a compact space \citep{wgan2017arjovsky}. In this setting, the discriminator is no longer trained to directly label samples as ``real'' or ``fake,'' but rather to learn the Lipschitz function needed to compute $W$. And, as the model training proceeds to minimize the loss function, the distance $W$ decreases, signifying that the generated output distribution is becoming closer to the real data distribution \citep{gulrajani2017WGANGP}.
Although more stable compared to GANs, the performance of WGAN-CPs was shown to be limited because: 1) small clipping thresholds lead to vanishing gradients while larger thresholds result in exploding gradients, and 2) the discriminator is biased to converge to simplified approximations of the Lipschitz function \citep{gulrajani2017WGANGP}. Improved training of WGANs was proposed by Gulrajani et al.\citep{gulrajani2017WGANGP}, who implement a gradient penalty method (WGAN-GP) instead of weight clipping to constrain the discriminator gradient. WGAN-GP enforces the Lipschitz constraint by imposing a penalty on the gradient norm if it is not close to the theoretical value of 1.}

\changes{
In this work, we test the performance of WGAN-CP and WGAN-GP trained with $1\text{,}000$ samples from our Cahn-Hilliard dataset. Using the PyTorch library \citep{paszke2017automatic}, we train typical convolutional feedforward neural networks for both the generator and the discriminator networks of WGAN-CP and WGAN-GP for a total of $23\text{,}690\text{,}498$ trainable parameters, $12\text{,}656\text{,}257$ for the generator network and $11\text{,}034\text{,}241$ for the discriminator network. We accomplish this using the code published in conjunction with \citep{githubWGAN} as a starting point. We perform no additional parameter tuning and keep all hyper-parameters at their default values.}

\subsubsection{\changes{Style-Based Generative Adversarial Networks}
\label{sec:stylgan}
}
A third approach to enhancing GANs involves modifying the latent space distributions of the generator network via feature mapping, and incorporating adaptive instance normalization (AdaIN) \citep{karras2019stylegan}. The AdaIN operation was first implemented by Huang and Belongie \citep{huang2017arbitrary} in style transfer algorithms \citep{gatys2016style}; transferring the style of one image to the content of another image. Specifically, AdaIN first normalizes each feature map and then scales its mean and variance according to a style input. 

In these StyleGAN models, the adjustments to the traditional generator are twofold: 1) the input latent space is mapped to a much less entangled intermediate latent feature space via an 8-layer multilayer perceptron network, and 2) the generator output is controlled by AdaIN processes which are themselves controlled by learned affine transformations that concentrate the intermediate latent space to specific styles that dictate the dominant image features at each convolution layer \citep{karras2019stylegan}. The StyleGAN2 architecture was later developed to remedy artifacts observed in StyleGAN generated images \citep{karras2020analyzing}. The StyleGAN2 using adaptive discriminator augmentation (StyleGAN2-ADA) \citep{karras2020training} is an adaptation of StyleGAN2 specifically designed for small training datasets. 
For the simplest implementations of training GANs with augmented datasets, generated distributions are known to exhibit features that are present in the augmented dataset, but not in the original dataset \citep{karras2020training, zhang2020consistency, zhao2020improved}. Therefore, to avoid this undesirable outcome, Karras et al. \citep{karras2020training} proposed the ADA method. 

For the augmentations to be ``non-leaking'' (i.e., not present in the generated examples) and for the GAN model to learn the true input distribution given an augmented dataset, the set of applied distortions for augmentation are required to be differentiable and belong to an invertible transformation of a probability distribution function \cite{karras2020training, Bora2018AmbientGANGM}. This can be achieved for a diverse set of possible augmentations when they are applied to the dataset with a probability $p$, with $0 < p < 0.8$ \citep{karras2020training}. However, the target value of $p$ is sensitive to the size of the dataset and as such, setting a fixed value for it is far from optimal. For this reason, Karras et al. \citep{karras2020training} implemented the discriminator augmentation method in an adaptive manner where $p$ is set to $0$ initially and its value is automatically adjusted (increased or decreased) based on a metric that indicates the extent by which the discriminator is overfitting. This heuristic is obtained from the discriminator outputs for the training and validation datasets, as well as the generated images and their mean over a fixed number of consecutive minibatches.
ADA can be implemented on any GAN model without modifying the network architecture or increasing training cost \citep{karras2020training}.
Notably, the StyleGAN2-ADA combination performs exceptionally well on the limited CIFAR-10 dataset \citep{krizhevsky2014cifar}, thus motivating our \changes{implementation} of the approach in this work. 

Here, we train the StyleGAN2-ADA model using the PyTorch library \citep{paszke2017automatic} with the code provided in \citep {karras2020training} on a small subset ($1\text{,}000$ samples) of our Cahn-Hilliard patterns. Of the set of transformations tested in \citep{karras2020training}, we apply the ones that contextually fit the Cahn-Hilliard dataset – geometric and color transformations. Geometric distortions include pixel blitting, isotropic and anisotropic scaling, fractional translation, and less frequently arbitrary rotation. \changes{We briefly note at this point that these distortions are implemented during the generation of synthetic patterns only and are not related to the equibiaxial loading conditions of the Finite Element simulations performed later once the generated data patterns are obtained.} For color transformations, the image brightness, contrast, and saturation were adjusted, the luma axis was flipped, and the hue axis was rotated arbitrarily. We perform no parameter tuning and keep all hyper-parameters at their default values. In total, the generator network has $22\text{,}238\text{,}990$ trainable parameters and the discriminator network has $23\text{,}406\text{,}849$ trainable parameters.

\subsection{\changes{Augmenting the Training Dataset with ``Procedural'' and ``Bernoulli'' Randomly Generated Patterns}
\label{sec:rbPatterns}}
\changes{As discussed in Section \ref{sec:meth_gen_model} and later depicted in Fig. \ref{fig:res_1} and Fig. \ref{fig:res_strainHist}, the three different ML-based generative models, WGAN-CP, WGAN-GP, and StyleGAN2-ADA are able to generate synthetic patterns relevant to the real Cahn-Hilliard patterns without being explicitly programmed to do so. 
However, there is a rich history of implementing ``Procedural'' algorithms for material microstructure pattern generation \citep{cule1999generating,fujii2001composite,jiao2007modeling,REDENBACH2009microstructure,PASKO2011Procedural,WALTERS2020Volumetric}. For example, many researchers have created explicitly programmed algorithm that draws from experimentally obtained probability distributions for creating and placing microstructural features within a domain \citep{VAUGHAN2010combined,ROMANOV2013Statistical}. These algorithms range from quite simple (e.g., Voronoi tessellation \citep{aurenhammer1991voronoi,FALCO2017Generation}) to quite complex (e.g., feature shape and placement based on energy minimization \citep{BARGMANN2018Generation}).
In this manuscript, we implement two additional pattern generation algorithms to compare to the ML-based generative models. 
First, we implement a straightforward ``Procedural'' algorithm where we create synthetic patterns with spatial correlations. In Section \ref{sec:res_disc}, we refer to these patterns as ``Procedural'' patterns. Second, we create random patterns following a ``Bernoulli'' distribution without spatial correlation. In Section \ref{sec:res_disc}, we refer to these patterns as ``Bernoulli'' patterns.} 

\changes{For the ``Procedural'' patterns, we begin with a low resolution grid, a $4\times4$, an $8\times8$, or a $16\times16$ grid, and assign each of the grid pixels an independent and identically distributed random value drawn from a uniform distribution $\mathcal{U}$[0,1]. Using the multidimensional image processing package in SciPy ``scipy.ndimage'' \citep{scipy}, we then increase the resolution of the resulting grayscale random image to the desired size of $64\times64$ and convert the upscaled image to a binary pattern by setting a brightness threshold.
For the ``Bernoulli'' patterns, we obtain binary images by simply creating a $64\times64$ grid of zeros, and then replacing the zeros with ones based on a probability threshold $p=0.6594$. For both types of patterns, the value of the threshold was chosen so that the light--to--dark ratio present in the real patterns is preserved. Notably, the ``Procedural'' patterns lead to spatially correlated features while the ``Bernoulli'' patterns do not.}

\subsection{\changes{Evaluation Metrics}
\label{sec:evalM}}

For evaluating and comparing the performance of the implemented GANs \changes{and the procedural methods} at creating generated examples, we considered three indicators. First, we compute the Fr\'{e}chet inception distance (FID) score, a quantitative metric to compare the resemblance between the distributions of the generated and real images \citep{heusel2017gans}. 
The FID, also known as Wasserstein-2 distance, is computed between the $2\text{,}048$ dimensional feature vectors, taken as the output of the last pooling layer of the pre-trained Inception network, of real and generated images by \citep{heusel2017gans}: 
\begin{equation}
\text{FID} = \left \|\bm{\mu_{1}} - \bm{\mu_{2}} \right\|^{2}_{2} + Tr\left[\mathbf{C_{1}}+\mathbf{C_{2}}-2\left(\mathbf{C_{1}} \mathbf{C_{2}}\right)^{1/2}\right]
\end{equation}
where $\bm{\mu_{1}}$ and $\bm{\mu_{2}}$, and $\mathbf{C_{1}}$ and $\mathbf{C_{2}}$ are the means and covariance matrices of the real and generated feature vectors respectively. The lower the FID score, the higher the similarity between the generated and the real images, with a $\text{FID} = 0$ indicating that the two sets are identical.
Second, we perform visual inspection of the generated patterns to check for the presence of any artifacts in the generated images and confirm their resemblance to real patterns. Finally, we perform an assessment of the diversity of the generated patterns by comparing the change in strain energy ($\Delta \Psi$) obtained from Finite Element simulations performed on the generated patterns to the same quantity obtained from simulations performed on real patterns from the Cahn-Hilliard dataset. The performance of our generative \changes{approaches} is reported in Section \ref{sec:res_generative}.

\subsection{\changes{Note on Standard Rotation-Based Augmentation}
\label{sec:rotation}
}
In addition to augmenting our training dataset with generated patterns, we further augment the training dataset of the metamodel by performing direct transformations on both real and generated input patterns. This type of straightforward data augmentation occurs after the real and generated input patterns have been used to run Finite Element simulations. Because we are considering an equibiaxial extension load case in this work, we can increase the size of the training dataset by a factor of $4$ by applying a set of predefined rotations ($0^{\circ}$, $90^{\circ}$, $180^{\circ}$, $270^{\circ}$) on the input images. For all $4$ rotated scenarios, the FEA simulation output $\Delta \Psi$ is identical, thus we can gain four data points per pattern. We report the significance of this standard augmentation on the metamodel performance in Section \ref{sec:res_transfer}.

\section{Results and Discussion}
\label{sec:res_disc}

In this Section, we report the results of employing the methods described in Sections \ref{sec:meth_simple_metamodel}, \changes{ \ref{sec:meth_gen_model}, \ref{sec:rbPatterns}, and \ref{sec:rotation}} to augment a small dataset of input patterns and train a convolutional neural network to predict the change in strain energy $\Delta \Psi$ for a given magnitude of applied equibiaxial extension. We begin in Section \ref{sec:res_generative} by describing the performance of the generative model\changes{s} when trained with just $1\text{,}000$ examples of real Cahn-Hilliard patterns. Then, in Section \ref{sec:res_gen_perform}, we demonstrate the performance of a metamodel where the training set contains simulations based on both real and generated input patterns. \changes{Finally, in Section \ref{sec:res_transfer}, we summarize the results of our transfer learning approach and the effect of standard rotation-based augmentations on metamodel performance.}

\subsection{Generative Model Performance} 
\label{sec:res_generative}

\begin{figure}[h]
\begin{center}
\includegraphics[width=.95\textwidth]{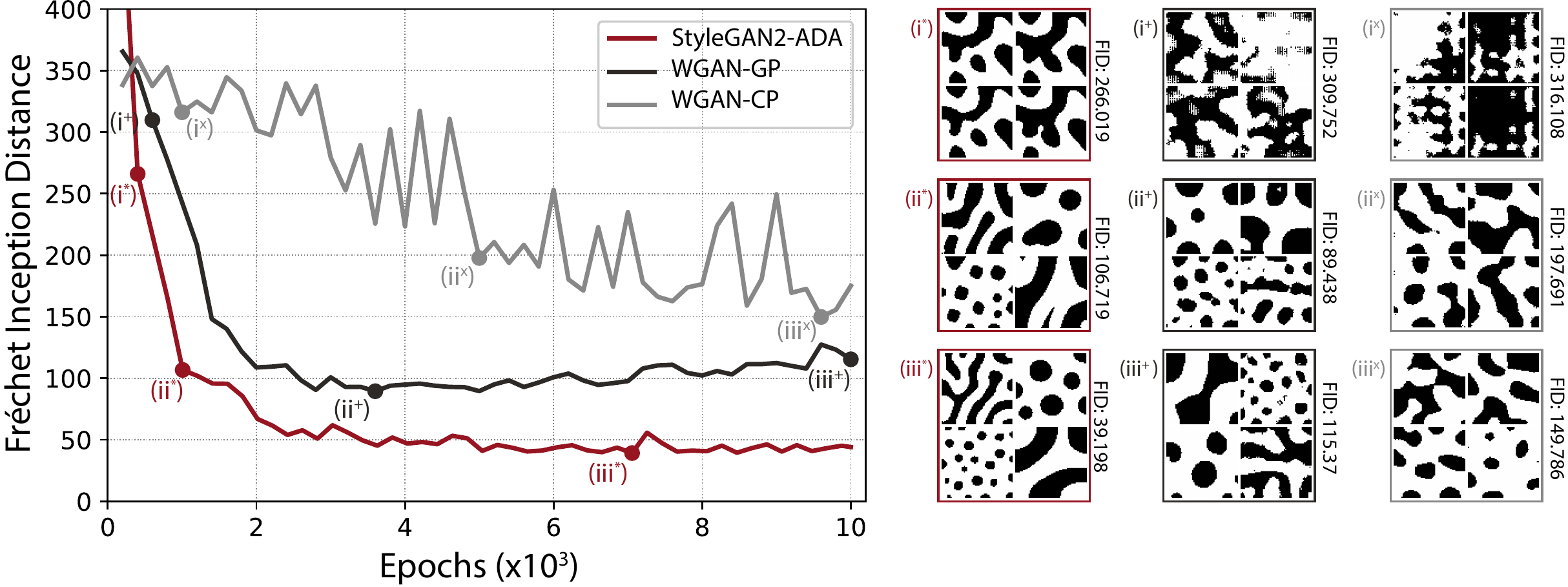}
\caption{\label{fig:res_1} \changes{Fr\'{e}chet Inception Distance (FID) with respect to the number of epochs for the StyleGAN2-ADA, WGAN-CP, and WGAN-GP ML-based generative models. In the right panel, we include examples of output patterns as model training proceeds to visualize the relation between a lower FID value and improved resemblance to the real input pattern. We note that all ML-based generative models are trained with just $1\text{,}000$ examples.}}
\end{center}
\end{figure}

As stated previously, we have tested three different GAN models, WGAN-CP, WGAN-GP, and SyleGAN2-ADA, with the aim of generating input patterns from a small training dataset of $1\text{,}000$ real Cahn-Hilliard patterns. In this Section, we show the performance of these methods and demonstrate that the StyleGAN2-ADA approach performs best at capturing the Cahn-Hilliard dataset. 
In Fig. \ref{fig:res_1}, we illustrate the performance by plotting the Fr\'{e}chet Inception Distance (FID) between $1\text{,}000$ real and $1\text{,}000$ generated patterns with respect to the number of epochs used for training. 
This plot shows that the StyleGAN2-ADA approach consistently has the lowest FID and is thus producing patterns that are a better match to the real dataset. We note that as expected, the calculated FID on real vs. real patterns converged to zero as we increased the size of the comparison datasets of patterns from $1\text{,}000$ (FID $\approx 13.3$) to $10\text{,}000$ (FID $\approx 1.7$).
In addition, we have annotated the plot in Fig. \ref{fig:res_1} with illustrated examples of generated patterns from \changes{the three generative models}. These illustrations not only confirm the intuition that as the FID decreases the patterns in the generated images more closely resemble those in the real dataset, but also show that for a converged model performance, the generated patterns look quite qualitatively realistic. Based on the higher FID for the WGAN-CP and WGAN-GP models, and the fact that FID begins to increase as the number of epochs increases, we conclude that both are inferior approaches \changes{when the goal is to generate realistic patterns that closely match the original dataset}. \changes{However, we note that in terms of model training time, the StyleGAN2-ADA network is significantly more expensive to train with the training process taking approximately $7.5$ hours on $4$ NVIDIA Tesla V100 GPUs. In comparison, it took approximately $0.5$ hours to train each of the WGAN-CP and WGAN-GP models on NVIDIA GeForce RTX 3060 Ti.}

\begin{figure}[p]
\begin{center}
\includegraphics[width=.5\textwidth]{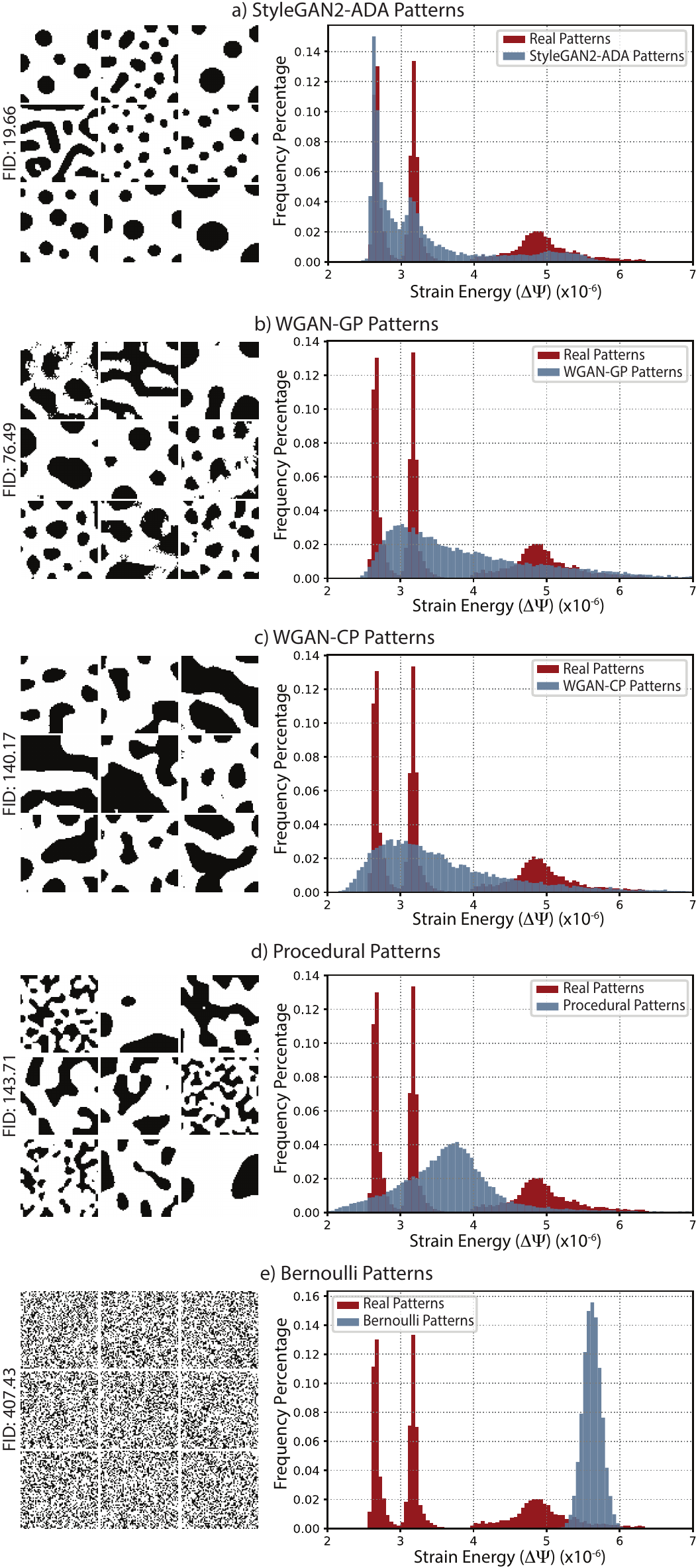}
\changes{\caption{\label{fig:res_strainHist} Visualization of the ML-based and procedural generative model results in order of increasing FID. For each pattern type, we show a comparison of strain energy $\Delta \Psi$ at $d=0.001$ for real and generated patterns with low fidelity data for: a) StyleGAN2-ADA patterns, b) WGAN-GP patterns, c) WGAN-CP patterns, d) ``Procedural'' patterns and, e) ``Bernoulli'' patterns. Overall, patterns produced with StyleGAN2-ADA have the highest similarity to the real dataset. We note that all ML-based generative models were trained with just $1\text{,}000$ examples, whereas the ``Procedural'' and ``Bernoulli'' patterns rely on no training data, only a knowledge of the average number of bright and dark pixels in the target dataset.}
}
\end{center}
\end{figure}

\changes{In Fig. \ref{fig:res_strainHist}, we plot the percentage frequency distribution of the change in strain energy $\Delta \Psi$ for $15\text{,}000$ low fidelity real and generated patterns subjected to small displacement ($d=0.001$) with equibiaxial extension Finite Element simulations. From comparing the distributions of $\Delta \Psi$, it appears that the StyleGAN2-ADA output distribution bears the most similarity to the real dataset. However, even though the WGAN and ``Procedural'' patterns are less authentic than StyleGAN2-ADA patterns, they are more divergent from the original $1\textit{,}000$ example real dataset while still maintaining overlap with the real distribution of $\Delta \Psi$. Lastly, the ``Bernoulli'' patterns appear only weakly relevant to the real dataset. From performing these simulations, we now have multiple datasets of low fidelity Finite Element simulations based on both real and generated input patterns that we can use to augment our ML model training datasets.}

\subsection{Metamodel Performance with an Augmented Training Dataset}
\label{sec:res_gen_perform}

\begin{figure}[h]
\begin{center}
\includegraphics[width=.8\textwidth]{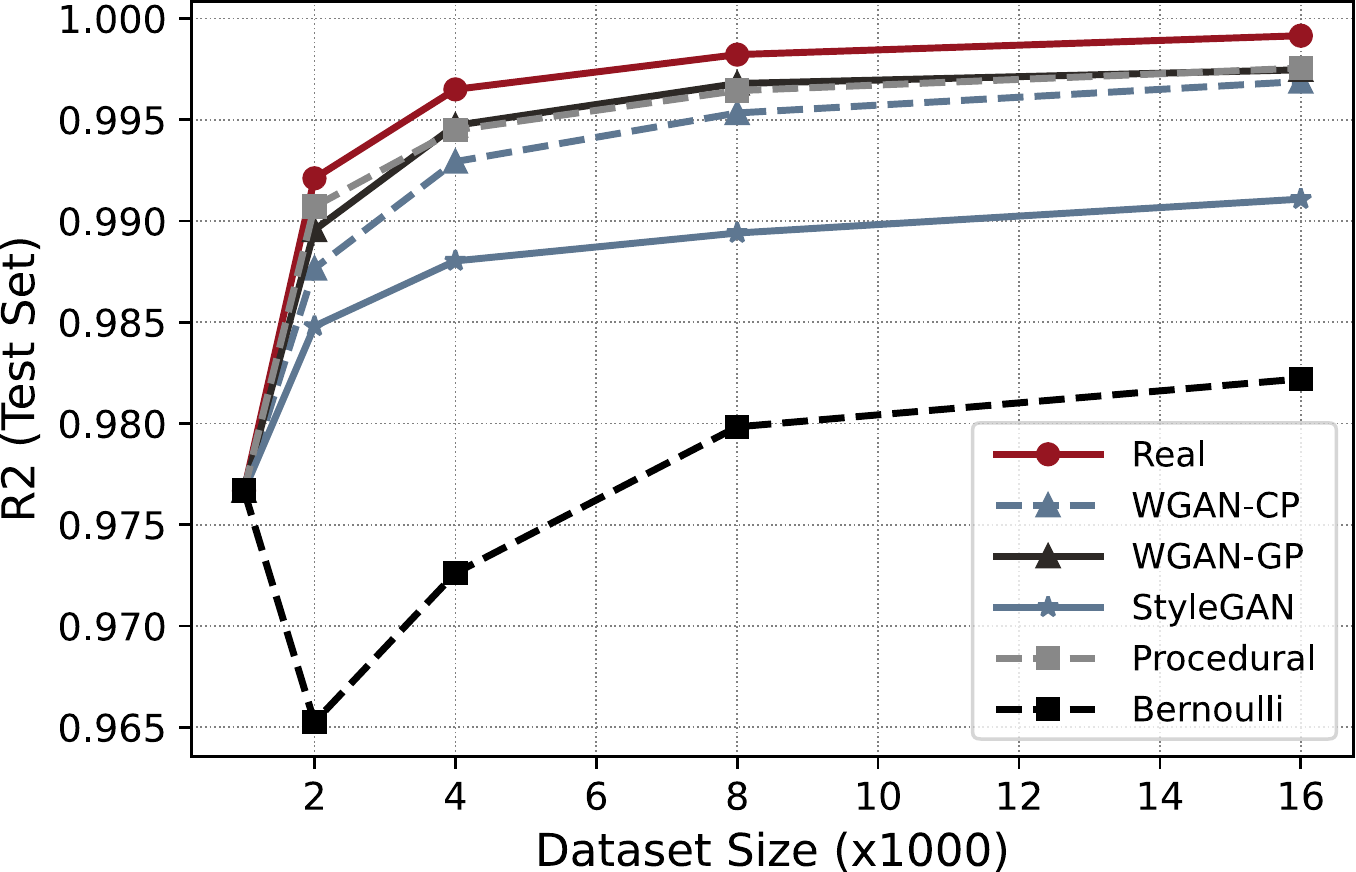}
\caption{\label{fig:res_2} Metamodel performance with respect to the size of the training dataset. Note that ``Dataset Size'' refers to the combined number of unique real and \changes{generated synthetic patterns}. For a dataset of \changes{$16\text{,}000$} real patterns, R2 is $0.9992$. For a dataset of $1\text{,}000$ real and \changes{$15\text{,}000$ synthetic patterns, the metamodel performance with ``Procedural'' and WGAN-GP patterns is almost identical with R2 values of $0.9975$. For augmentations with WGAN-CP, StyleGAN2-ADA, and ``Bernoulli'' patterns, the corresponding R2 values are $0.9969$, $0.9911$, and $0.9822$ respectively.}}
\end{center}
\end{figure}

With our trained \changes{ML-based generative models and procedural algorithm-based generative models}, we are able to generate synthetic input patterns and use them as inputs to Finite Element simulations where the results are used to augment our metamodel training datasets. In Fig. \ref{fig:res_2}, we show the test performance of the CNN-based metamodel defined in Section \ref{sec:basic_CNN} trained on these data. We report the R2 score computed on held out test data with respect to dataset size for five different types of training dataset. The first training dataset type is composed of real patterns only. The rest of the training dataset types contain a fixed number of real data points ($1\text{,}000$), and the size of \changes{these datasets} is increased by adding simulation results obtained from patterns \changes{generated using WGAN-CP, WGAN-GP, StyleGAN2-ADA, ``Procedural'', or ``Bernoulli'' methods, respectively}. 
For all training dataset types, we consider sample sizes of $1\text{,}000$, $2\text{,}000$, $4\text{,}000$, $8\text{,}000$, and $16\text{,}000$ patterns. \changes{For reference, training our CNN-based metamodel for $100$ epochs with $16\text{,}000$ samples took $\approx 2$ minutes on a single Nvidia Tesla V100 GPU. The results presented in Fig. \ref{fig:res_2} reveal that metamodels trained with WGAN-GP and ``Procedural'' patterns perform nearly equivalently with R2 scores of $0.9975$, and exhibit only a slightly inferior performance to a metamodel trained entirely on real patterns (R2 $= 0.9992$)} for a dataset size of $16\text{,}000$. Notably, in all cases, the addition of the generated input patterns improves the performance of the metamodel \changes{except for $1\text{,}000$ and $3\text{,}000$ random ``Bernoulli'' patterns which decreased the metamodel performance. This is anticipated because these patterns are not spatially correlated the way real Cahn-Hilliard patterns are, as depicted in Fig. \ref{fig:res_strainHist}. In fact, we find the very slight improvement in the performance of the metamodels when the dataset is augmented with more than $8\text{,}000$ of this type of synthetic patterns to be counter intuitive. Comparing the metamodel performance for dataset augmentations with StyleGAN2-ADA patterns versus WGAN-GP and ``Procedural'' patterns, we anticipate that the diversity of the WGAN-GP and ``Procedural'' synthetic patterns proves to be more important than the authenticity of the StyleGAN2-ADA patterns for enhancing metamodel performance. Namely, even though the \changes{StyleGAN2-ADA} patterns were closer to real patterns than the WGAN-GP patterns, they were perhaps less diverse or even too similar to the real patterns used for training and thus less beneficial when training the predictive model.}

\subsection{\changes{Metamodel Performance with Transfer Learning}
\label{sec:res_transfer}}

After training the metamodels on \changes{datasets} based on low fidelity simulation data, we evaluate the efficacy of our straightforward transfer learning approach described in Section \ref{sec:tl} to make predictions on the corresponding high fidelity simulation dataset. We begin with our metamodel pre-trained using the weights obtained from our low fidelity dataset metamodel trained with $1\text{,}000$ real and $15\text{,}000$ generated patterns with rotation-based augmentation \changes{as described in Section \ref{sec:rotation}. With this additional rotation-based augmentation, a dataset size of $N$ corresponds to $4N$ training points.} Then, we perform additional training with $1\text{,}000$ real pattern-based high fidelity simulations. \changes{As shown in Table \ref{table:res_tf}, this} transfer learning-based training process predicts the change in strain energy $\Delta \Psi$ at the final displacement for test data with \changes{R2 score of $0.9977$ and corresponding MAE of $0.0074$ when the weights are initialized with the best performing metamodel trained with a dataset augmented with StyleGAN2-ADA patterns in addition to the rotation-based methods. We note that although the performance of metamodels trained with datasets augmented with WGAN-CP, WGAN-GP and ``Procedural'' patterns (with or without additional rotations), is better than equivalent metamodels trained based on StyleGAN2-ADA augmented datasets (see Fig. \ref{fig:res_2} and Table \ref{table:res_tf} ``Pre-Training'' column), the StyleGAN2-ADA augmented model performs best after transfer learning.} Alternatively, training a metamodel initialized with random weights predicts $\Delta \Psi$ for the same high fidelity dataset with an R2 of \changes{$0.9783$} and corresponding MAE of \changes{$0.0198$}. We note that the real patterns used in the training and test sets in the low fidelity metamodel training match the patterns used in the high fidelity metamodel training, and that the same $1\text{,}000$ real patterns are used as training data for our metamodels and the generative model\changes{s}. Overall, this demonstrates that \changes{synthetic pattern-based} and rotation-based data augmentation strategies can be combined with our previously explored transfer learning approach \citep{lejeune2021exploring} to create meaningful training datasets that rely on only a small number of representative input pattern images and are computationally cheap to generate. \changes{Based on our investigations, we find that ``Procedural'' patterns, when possible to generate, can not only be an effective choice, but also may be a better choice than ML-based generative models in some circumstances. When it is not possible to generate procedural patterns, our results indicate that \changes{both} WGAN-GP and StyleGAN2-ADA are good choices for ML-based generative models.}

\begin{table}[h]
  \centering
  \caption{\label{table:res_tf}\changes{Results of transfer learning. Pre-training is performed with low fidelity data of $1\text{,}000$ real Cahn-Hilliard patterns and $15\text{,}000$ either real or generated patterns subjected to $3$ additional rotations of the unique domains. Fine-tuning refers to training a metamodel with $1\text{,}000$ high fidelity real Cahn-Hilliard patterns with the model initial weights ``transferred'' from the pre-trained model on the corresponding row. For a metamodel that is trained on $1\text{,}000$ entirely real Cahn-Hilliard patterns without transfer learning, the model weights are randomly initialized. Overall, it is evident that our transfer learning approach improves the MAE by at least $55\%$ when predicting change in strain energy. Representative plots of true strain energy vs. predicted strain energy are shown in Appendix \ref{appendix-a}, Fig. \ref{fig:r2-eval} to add additional context to these values.}}
    \resizebox{0.75\textwidth}{!}{\begin{tabular}{ccccc}
    \toprule
    \multicolumn{5}{c}{\textbf{Transfer Learning}}\\
    \midrule
    \multicolumn{2}{c}{\textbf{Pre-Training (Low Fidelity)}} & & \multicolumn{2}{c}{\textbf{Fine-Tuning (High Fidelity)}}\\
    Pattern Generation Method & R2 Score (Test Set) & & R2 Score (Test Set) & MAE (Test Set)\\
    \midrule
    Real & 0.9991 & & 0.9991 & 0.0046\\
    StyleGAN2-ADA & 0.9967 & & 0.9977 & 0.0074\\
    WGAN-CP & 0.9973 & & 0.9974 & 0.0088\\
    WGAN-GP & 0.9981 & & 0.9974 & 0.0077\\
    ``Procedural'' & 0.9979 & & 0.9973 & 0.0084\\
    \midrule
    \multicolumn{2}{c}{\textbf{No Transfer Learning}} & & 0.9783 & 0.0198\\
    \bottomrule
    \end{tabular}%
  \label{tab:addlabel}}%
\end{table}%

\section{Conclusion}
\label{sec:conc}

In this paper, we extend our previous work on using machine learning-based metamodels to predict mechanical quantities of interest in heterogeneous materials \citep{lejeune2021exploring,lejeune2020mechanical, mohammadzadeh2022predicting} to include a method for working with size-limited datasets. Specifically, we are interested in developing tools for making smaller datasets (with as few as $1\text{,}000$ example input patterns) amenable to deep learning approaches. To accomplish this, we first create a new dataset of spatially heterogeneous domains undergoing large deformation with material property patterns based on the Cahn-Hilliard equation, the ``Mechanical MNIST Cahn-Hilliard'' dataset. In contrast to our previous work \citep{lejeune2020mechanicalF,lejeune2019mechanicalUE}, these input patterns are more relevant to heterogeneous biological materials. In this paper, we present a brief overview of the underlying theory behind the Cahn-Hilliard equations, and describe the procedure for generating the dataset. Then, with this dataset, we test the efficacy of different Generative Adversarial Network (GAN) models at generating new Cahn-Hilliard patterns from a limited training dataset of $1\text{,}000$ example patterns. Of the approaches that we explored, we found that the StyleGAN2-ADA model performed best at generating synthetic Cahn-Hilliard patterns \changes{(FID $=39.2$)}. \changes{In addition to GAN-based synthetic patterns, we explored two procedural approaches and created two additional types of synthetic Cahn-Hilliard patterns, ``Procedural'' patterns and spatially uncorrelated ``Bernoulli'' patterns.} With \changes{ML-based and procedural-based} generated patterns, we then created low fidelity (i.e., computationally cheap through coarse mesh and perturbation displacements) Finite Element simulation \changes{datasets} comprised of $1\text{,}000$ simulations based on real input patterns and \changes{$15\text{,}000$} simulations based on generated patterns. We then compared the performance of metamodel\changes{s} trained on \changes{these} hybrid real and generated input pattern dataset\changes{s} to a metamodel trained entirely on real patterns and found that our data augmentation approaches were highly effective (R2 of \changes{$0.9975$ for ``Procedural'' and WGAN-GP augmentation-based datasets and R2 of $0.9992$} for the dataset based entirely on real patterns). In addition, we built on our previous work in using transfer learning to leverage low fidelity simulation datasets \citep{lejeune2021exploring}, and demonstrated that with just $1\text{,}000$ high fidelity (i.e., refined mesh, full applied displacement) Finite Element simulations, we could transfer a low fidelity metamodel to the high fidelity dataset \changes{and obtain an R2 score of $0.9976$ and corresponding MAE of $0.0074$} for predicting change in strain energy. This final result was obtained with $1\text{,}000$ unique real input patterns, $1\text{,}000$ real pattern low fidelity simulations, $1\text{,}000$ real pattern high fidelity simulations, and \changes{$15\text{,}000$ low fidelity simulations with StyleGAN2-ADA generated input patterns}. 

Broadly speaking, we anticipate that the work presented in this paper will motivate multiple future research directions. To this end, we have made both our ``Mechanical MNIST Cahn-Hilliard'' dataset and our metamodel implementation readily available for other research groups to build on under open-source licenses (see Section \ref{sec:additional_info}). In the future, we anticipate that others may implement alternative approaches to this problem that exceed the baseline performance established in this paper. Here, we established baseline performance for \changes{three} problems: (1) training generative model\changes{s} with just $1\text{,}000$ example patterns, (2) \changes{demonstrating the effectiveness of simple procedural data generation and augmentation approaches,} and (3) training a metamodel based on a Finite Element simulation dataset where the relevant material property distribution is defined by just $1\text{,}000$ example patterns. However, because our dataset is published under an open source license, others are free to formulate different challenges and attempt the same problem with an entirely different metamodeling approach. In particular, we anticipate future work in developing more sophisticated approaches for representing the input pattern space, future work in predicting full field quantities of interest in addition to the single quantity of interest predicted here, \changes{and future work in accounting for more aspects that render modeling soft tissue very challenging, such as material anisotropy and the broad uncertainty in material properties.} In addition, we plan to extend the ``Mechanical MNIST Cahn-Hilliard'' dataset to include additional constitutive parameters, a more diverse set of constitutive models, and additional loading scenarios in the future. \changes{In addition, we note that there should be further future investigation into the minimum number of data points required to train a GAN model for this type of problem. In this work, we relied entirely on a pragmatic selection of $1\text{,}000$ data points simply because $100$ would likely be insufficient for training a GAN, and $10\text{,}000$ would no longer be resolutely in the size-limited datasets regime.} Looking forward, we hope that the findings in this work will make deep learning-based metamodels much more accessible for researchers working with limited examples of their input pattern spaces of interest.  

\section{Additional Information}
\label{sec:additional_info}

The ``Mechanical MNIST Cahn-Hilliard'' dataset is available through the OpenBU Institutional Repository at \url{https://open.bu.edu/handle/2144/43971} \citep{MMistCH}. We provide with this dataset a supplementary document that includes more details on the theory of the Cahn-Hilliard equation and our Finite Element implementation.
The codes for generating the Cahn-Hilliard patterns and for performing the Finite Element equibiaxial extension simulations using FEniCS computing platform (\url{https://fenicsproject.org})
are available on github at \url{https://github.com/elejeune11/Mechanical-MNIST-Cahn-Hilliard}.
The codes for implementing the metamodel pipeline including the convolutional neural network model for a single quantity of interest prediction and the GAN model for data synthesis are also made available at \url{https://github.com/saeedmhz/cahn-hilliard}.

\section{Acknowledgements}
\label{sec:ack}
We would like to thank the staff of the Boston University Research Computing Services and the OpenBU Institutional Repository (in particular Eleni Castro) for their invaluable assistance with generating and disseminating the ``Mechanical MNIST Cahn-Hilliard Dataset''. This work was made possible through start up funds from the Boston University Department of Mechanical Engineering, the David R. Dalton Career Development Professorship, the Hariri Institute Junior Faculty Fellowship, the National Science Foundation Engineering Research Center CELL-MET NSF EEC-1647837, and the Office of Naval Research Award N00014-22-1-2066.

\appendix
\label{sec:appendix}
\section{\changes{Qualitative Visualization of Metamodel Performance}}
\label{appendix-a}

\begin{figure}[h]
\begin{center}
\includegraphics[width=.95\textwidth]{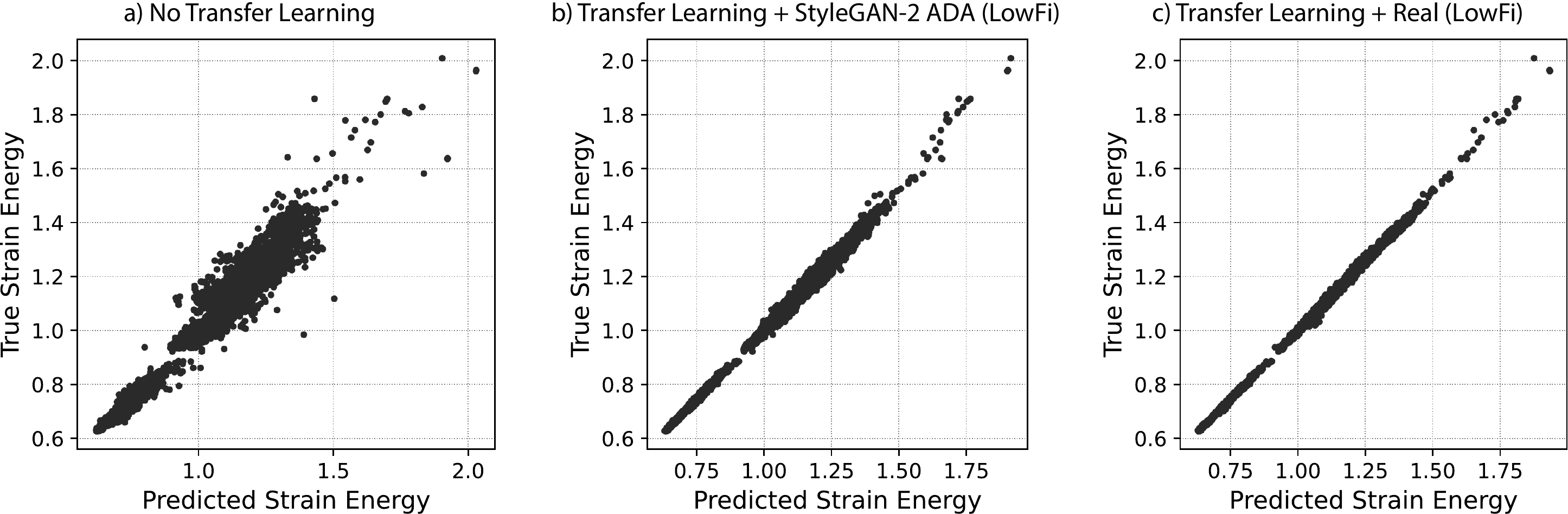}
\caption{\label{fig:r2-eval}\changes{Qualitative interpretation of R2 scores for transfer learning evaluation. True versus predicted strain energy values of high fidelity test data are plotted for three different metamodels trained with $1\text{,}000$ high fidelity real data points. (a) Metamodel weights are initialized randomly (i.e., no transfer learning is performed). (b) Metamodel weights are transferred from a model trained on $1\text{,}000$ low fidelity real samples and $15\text{,}000$ generated samples from StyleGAN2-ADA with extra rotation-based augmentations. (c) Metamodel weights are initialized by transferring weights of a model trained on $16\text{,}000$ low fidelity real data with extra rotation-based augmentations.}}
\end{center}
\end{figure}

\changes{In this Appendix, we provide additional information to support how synthetic data augmentation combined with a simple transfer learning approach can help improve the performance of our metamodel. As shown in Section \ref{sec:res_transfer}, initializing the weights of our metamodel with the weights of a model trained on low fidelity real data augmented with the proper set of generated data can increase the R2 score of predicted high fidelity strain energy values from $0.9783$ to $0.9977$. In order to qualitatively interpret the benefits of this improvement, we plotted true versus predicted strain energy values for all samples in the test set of the high fidelity dataset for three different models (Fig. \ref{fig:r2-eval}). Figure \ref{fig:r2-eval}a shows the results where no transfer learning is performed, whereas Fig. \ref{fig:r2-eval}b shows the case where the weights are transferred from a model trained on $1\text{,}000$ low fidelity real samples and $15\text{,}000$ synthetic samples generated from StyleGAN2-ADA with extra rotation-based augmentations. In Fig. \ref{fig:r2-eval}c the initial weights are transferred from a low fidelity model trained on $16\text{,}000$ real data with rotation-based augmentations. Overall, this figure further supports our findings from Table \ref{table:res_tf} and Section \ref{sec:res_transfer} where we state the performance of our metamodels in terms of R2 score.
}

\FloatBarrier

\bibliographystyle{unsrt}
\bibliography{references}  

\end{document}